\documentclass[10pt,twocolumn,letterpaper]{article}

\usepackage{cvpr}              

\usepackage[dvipsnames]{xcolor}
\usepackage{amsmath}
\usepackage{amssymb}
\usepackage{booktabs}
\usepackage{amsmath}
\usepackage{amssymb}
\usepackage{comment}
\usepackage{bm}
\usepackage{bbm}
\usepackage{algpseudocode}
\usepackage{microtype}      %
\usepackage{makecell}
\usepackage{mathtools}
\usepackage{multirow}
\usepackage{multicol}
\usepackage{diagbox}
\usepackage{pifont}
\usepackage{makecell} 
\usepackage{threeparttable,booktabs}

\newcommand{\method}{\textsc{SADM}\xspace}

\usepackage[ruled,vlined]{algorithm2e}
\usepackage{amsmath,amsfonts,bm}

\newcommand{\xmark}{\ding{55}}%
\newcommand{\cmark}{\ding{51}}%

\def\eqref#1{equation~\ref{#1}}

\def\1{\bm{1}}

\ifx\theorem\undefined
\newtheorem{theorem}{Theorem}
\fi
\ifx\example\undefined
\newtheorem{example}{Example}
\fi
\ifx\property\undefined
\newtheorem{property}{Property}
\fi
\ifx\lemma\undefined
\newtheorem{lemma}[theorem]{Lemma}
\fi
\ifx\proposition\undefined
\newtheorem{proposition}[theorem]{Proposition}
\fi
\ifx\remark\undefined
\newtheorem{remark}[theorem]{Remark}
\fi
\ifx\corollary\undefined
\newtheorem{corollary}[theorem]{Corollary}
\fi
\ifx\definition\undefined
\newtheorem{definition}[theorem]{Definition}
\fi
\ifx\conjecture\undefined
\newtheorem{conjecture}[theorem]{Conjecture}
\fi
\ifx\fact\undefined
\newtheorem{fact}[theorem]{Fact}
\fi
\ifx\claim\undefined
\newtheorem{claim}[theorem]{Claim}
\fi
\ifx\assum\undefined

\fi
\ifx\assum\undefined

\fi

\newcommand{\Eb}[2]{\E_{#1}\!\left[#2\right]}

\newcommand{\bI}{\mathbf{I}}

\newcommand{\bx}{\mathbf{x}}
\newcommand{\by}{\mathbf{y}}

\newcommand{\bepsilon}{{\boldsymbol{\epsilon}}}

\def\mI{{\bm{I}}}

\DeclareMathAlphabet{\mathsfit}{\encodingdefault}{\sfdefault}{m}{sl}
\SetMathAlphabet{\mathsfit}{bold}{\encodingdefault}{\sfdefault}{bx}{n}

\newcommand{\E}{\mathbb{E}}

\definecolor{cvprblue}{rgb}{0.21,0.49,0.74}
\usepackage[pagebackref,breaklinks,colorlinks,citecolor=cvprblue]{hyperref}

\def\paperID{4676} 
\def\confName{CVPR}
\def\confYear{2024}

\title{Structure-Guided Adversarial Training of Diffusion Models}

\author{Ling Yang$^{*\dag}$\quad Haotian Qian\thanks{Contributed equally.}\quad Zhilong Zhang\quad Jingwei Liu\quad Bin Cui\thanks{Corresponding authors.}\\
Peking University\\
{\tt\small yangling0818@163.com, zzdmqht@gmail.com, \{zzl2018math, bin.cui\}@pku.edu}
}

\begin{document}
\maketitle
\begin{abstract}
Diffusion models have demonstrated exceptional efficacy in various generative applications. While existing models focus on minimizing a weighted sum of denoising score matching losses for data distribution modeling, their training primarily emphasizes instance-level optimization, overlooking valuable structural information within each mini-batch, indicative of pair-wise relationships among samples. To address this limitation, we introduce \textbf{S}tructure-guided \textbf{A}dversarial training of \textbf{D}iffusion \textbf{M}odels (\textbf{SADM}). In this pioneering approach, we compel the model to learn manifold structures between samples in each training batch. To ensure the model captures authentic manifold structures in the data distribution, we advocate adversarial training of the diffusion generator against a novel \textit{structure discriminator} in a minimax game, distinguishing real manifold structures from the generated ones. SADM substantially improves existing diffusion transformers and outperforms existing methods in image generation and cross-domain fine-tuning tasks across 12 datasets, establishing a new state-of-the-art FID of \textbf{1.58} and \textbf{2.11} on ImageNet for class-conditional image generation at resolutions of 256$\times$256 and 512$\times$512, respectively. 
\end{abstract}
\section{Introduction}
Diffusion models \citep{sohl2015deep,ho2020denoising,song2020score,song2019generative,song2020improved,yang2023diffusion} have
achieved remarkable generation quality in various tasks, including image generation \citep{dhariwal2021diffusion,rombach2022high,ramesh2022hierarchical,saharia2022photorealistic,wang2022diffusion,yang2024improving,yang2024mastering,zhang2024realcompo},
audio synthesis \citep{chen2020wavegrad,kong2020diffwave,popov2021grad}, and interdisciplinary applications \citep{hoogeboom2022equivariant,xu2021geodiff,huang2024proteinligand}. 
Starting from tractable noise distribution, diffusion models generate data by progressively removing noise. This involves the model learning to reverse a pre-defined diffusion process that sequentially introduces varying levels of noise to the data. The model is parameterized and undergoes training by optimizing the weighted sum of denoising score matching losses \citep{ho2020denoising} for various noise levels \citep{song2019generative}, aiming to learn the recovery of clean images from corrupted images.
\begin{figure}[t]
\vspace{-3mm}
\begin{center}\centerline{\includegraphics[width=0.9\linewidth]{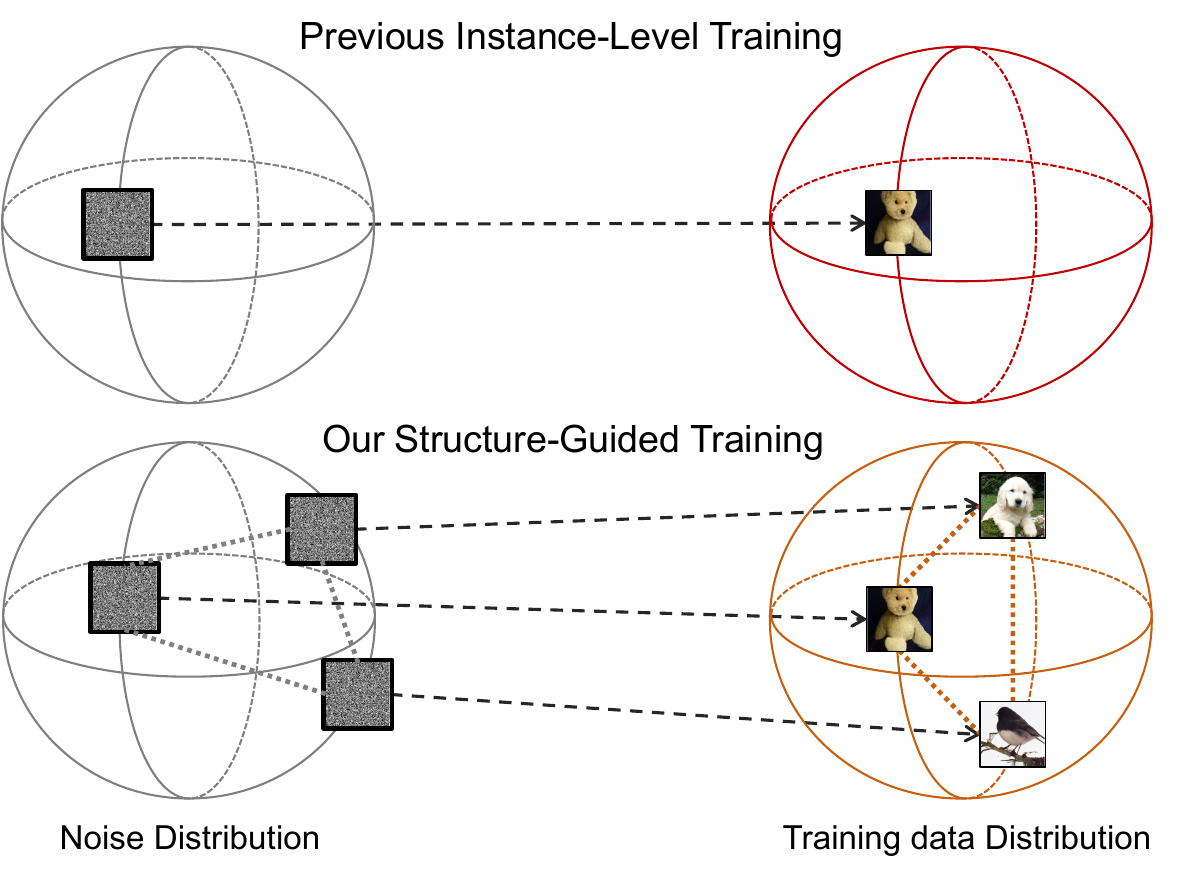}}
\vspace{-3mm}
\caption{Comparison between previous instance-level training and our structure-guided training for diffusion models.}
\label{pic-pre}
\end{center}
\vspace{-10mm}
\end{figure}

Aiming to maximally model the data distribution, recent works \citep{song2021maximum,lu2022maximum,vahdat2021score,kim2022refining,xu2023pfgmpp,yang2024crossmodal} attempt to improve the precision of training diffusion models. 
For instance, many approaches design effective weighting schemes \citep{song2021maximum,lu2022maximum,yang2024crossmodal,choi2022perception,kim2022soft,hang2023efficient} for maximum likelihood training or add an extra regularization term \citep{lai2022regularizing,pmlr-v202-ning23a,daras2023consistent} to the denoising score loss. There are also some works to enhance the model expressiveness by incorporating other generative models, such as VAE \citep{vahdat2021score,kingma2021variational,rombach2022high}, GAN \citep{xiao2021tackling}, and Normalizing Flows \citep{zhang2021diffusion,kim2022maximum,luo2021diffusion}.
However, their improved training of diffusion models primarily concentrates on instance-level optimization, overlooking the valuable structural information among batch samples.
This oversight is significant, as the incorporation of structural details is crucial for aligning the learned distribution with the underlying data distribution.

\begin{figure*}[ht]
\vspace{-2mm}
\begin{center}\centerline{\includegraphics[width=1\linewidth]{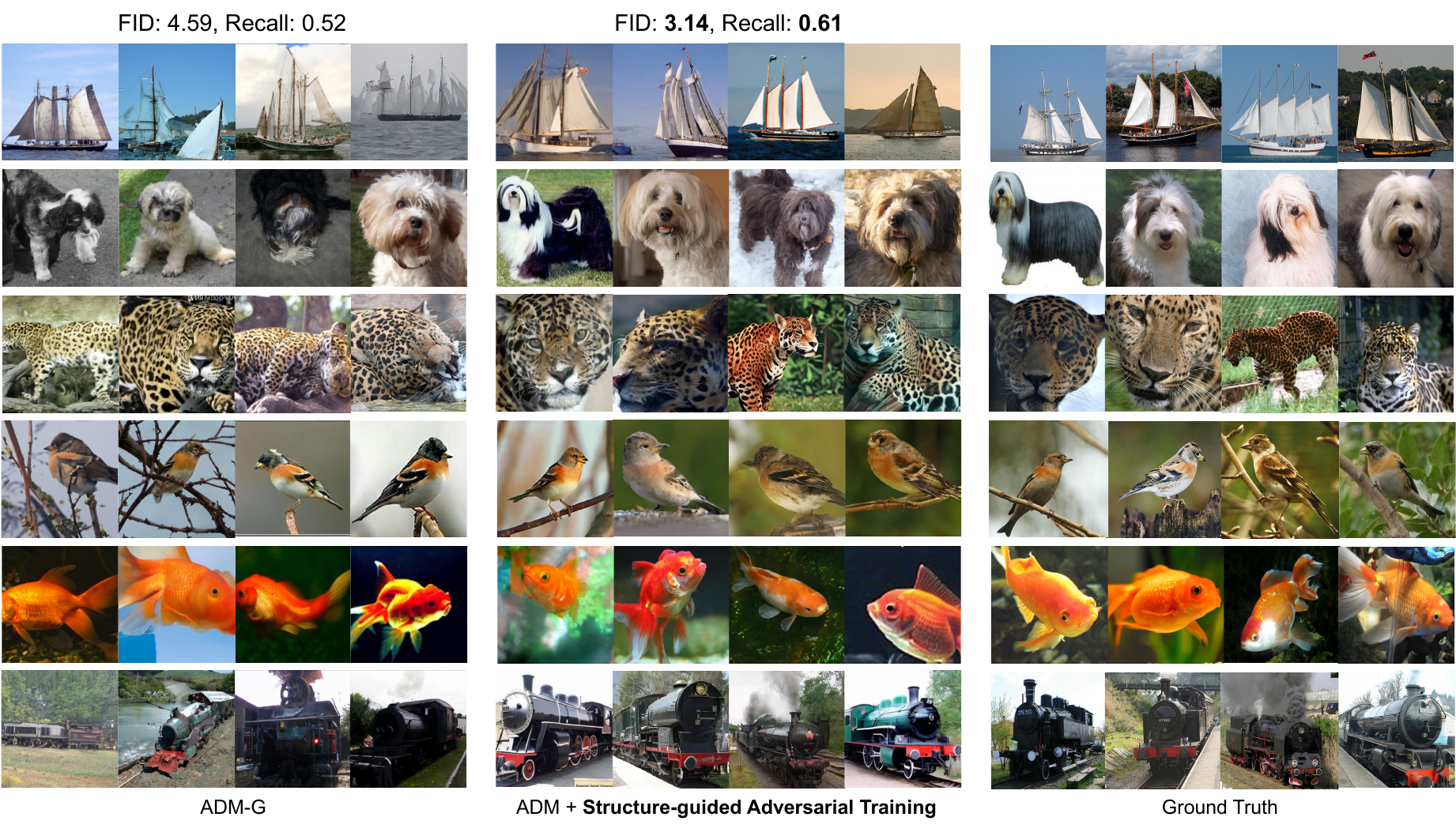}}
\vspace{-3mm}
\caption{Generated samples on ImageNet $256\times 256$ with (i) ADM with Classifier Guidance (ADM-G) \citep{dhariwal2021diffusion}, (ii) ADM optimized by our \textbf{Structure-guided Adversarial Training}, and (iii) real samples in ground truth classes. We can significantly improve diffusion models qualitatively and quantitatively, and our generated sample distribution is overally more similar to real sample distribution. See \cref{app-samples} for more synthesis samples of our SOTA model.}
\label{pic-result1}
\end{center}
\vspace{-5mm}
\end{figure*}

To mitigate the challenges posed by existing instance-level training methods, we propose \textbf{S}tructure-guided \textbf{A}dversarial training of \textbf{D}iffusion \textbf{M}odels (\textbf{SADM}). 
In contrast to conventional instance-level training illustrated in \cref{pic-pre}, our approach guides diffusion training at a structural level. During batch training, the model is facilitated to learn manifold structures within batch samples, represented by pair-wise relationships in a low-dimensional feature space. To accurately learn real manifold structures in the data distribution, we introduce a novel \textit{structure discriminator} that distinguishes genuine manifold structures from generated ones. For clarity, we alternatively refer to the diffusion models optimized through our structure-guided training as \textit{joint sample diffusion}, a concept theoretically proven to enhance diffusion model optimization.

We assess the performance of our model across two pivotal tasks: image generation and cross-domain fine-tuning. The former involves training the diffusion model from scratch to evaluate its capability in capturing the entire data distribution. The latter leverages a pre-trained diffusion model, constructed on a large-scale source dataset, and fine-tunes it on a target dataset to assess transferability. Our extensive experiments consistently demonstrate that our approach significantly improves the model's ability to effectively capture the underlying data distribution (\cref{pic-result1}). \method achieves state-of-the-art results across 12 image datasets, including ImageNet \citep{deng2009imagenet}. Furthermore, we observe its potential for facilitating rapid adaptation to new domains in cross-domain fine-tuning tasks.
We summarize our contributions as follows:
\begin{itemize}
    \item To the best of our knowledge, we are the first to propose \textbf{Structure-guided Adversarial Training} to optimize diffusion models from a structural perspective.
    \item We theoretically show that \method is superior in capturing real data distribution, and can generalize to various image- and latent-based diffusion architectures (e.g., DiT \citep{peebles2023scalable}). 
    \item We substantially outperform existing methods on image generation and cross-domain fine-tuning tasks, achieving a new \textbf{state-of-the-art FID of 1.58 and 2.11} on ImageNet at resolutions of 256$\times$256 and 512$\times$512, respectively.
\end{itemize}

\section{Related Work}
In this work, we focus on \textbf{improving the training of diffusion models}. Here, we review previous related works and compare our \method with them. 

\paragraph{Modifying Training Objectives of Diffusion Models} A line of research modifies training objectives to achieve state-of-the-art likelihood \citep{song2021maximum,lu2022maximum,yang2024improving,choi2022perception}.
\citet{song2021maximum} propose likelihood weighting to enable approximate maximum likelihood training of score-based diffusion models \citep{song2020score,song2019generative,song2020improved} while ContextDiff \citep{yang2024crossmodal} introduces an effective shifting scheme
for facilitating the diffusion and training processes of diffusion probabilistic models \citep{sohl2015deep,ho2020denoising} to achieve
improved sample quality with stable training.
\citet{lai2022regularizing} and \citep{daras2023consistent} introduce an extra regularization term to
the denoising score loss to satisfy some properties of the diffusion process. However, these improvements mainly focus on sample-level optimization, neglecting the rich structural information within batch samples, which is critical for aligning the learned distribution and data distribution.
Hence, we enforce the model to maximally learn the manifold structures of samples.

\paragraph{Combining Additional Models for Diffusion Training} Another line of research incorporates other models to improve the stability and precision of diffusion training. For example,
INDM \citep{kim2022maximum} expands the linear diffusion to trainable nonlinear diffusion through a normalizing flow to improve the training curve of diffusion models.
\citet{jolicoeur2020adversarial,kim2022refining} improve diffusion models with adversarial learning while \citet{xiao2021tackling} model each denoising step using a multimodal conditional GAN. LSGM \citep{vahdat2021score} and LDM \citep{rombach2022high} conduct diffusion process in the semantic latent space obtained with a pre-trained VAE.  
Although these combinations strengthen the model expressiveness for capturing data distribution, they are still limited in modeling the underlying manifold structures within training samples. We propose a novel \textit{structure discriminator} for adversarially learn the diffusion model from a structural perspective.

\section{Preliminary}
\label{sec:background}
\paragraph{Diffusion Models} We consider diffusion models~\citep{sohl2015deep,song2019generative,ho2020denoising} specified in continuous time~\citep{tzen2019neural,song2020score,chen2020wavegrad,kingma2021variational}.  Given samples $\bx_0$ from a data distribution $q_0(\bx_0)$, noise scheduling functions $\alpha_t, \sigma_t$, a diffusion model has latent variables $\bx = \{\bx_t \,|\, t \in [0,1]\}$, and the forward process is defined with $q(\bx_t|\bx_0)$, a Gaussian process satisfying the following Markovian structure:
\begin{align}
    q(\bx_t|\bx_0) &= \mathcal{N}(\bx_t; \alpha_t \bx_0, \sigma_t^2 \bI), \\\label{noising}
    q(\bx_t | \bx_s) &= \mathcal{N}(\bx_t; (\alpha_t/\alpha_s)\bx_s, \sigma_{t|s}^2\bI)
\end{align}
where $0 \leq s < t \leq 1$ and $\sigma^2_{t|s} = (1-e^{\lambda_t-\lambda_s})\sigma_t^2$, and $\lambda_t = \log[\alpha_t^2/\sigma_t^2]$ denotes the log signal-to-noise-ratio \citep{kingma2021variational}. The goal of the diffusion model is to denoise $\bx_t\sim q(\bx_t|\bx_0)$ by estimating $\hat\bx_\theta(\bx_t) \approx \bx_0$. We train this denoising model $\hat\bx_\theta$ using a weighted mean squared error loss
\begin{align}
    \Eb{\bx_0,\bepsilon,t}{w(\lambda_t) \|\hat\bx_\theta(\bx_t) - \bx_0 \|^2_2} \label{denosing-object-1}
\end{align}
over uniformly sampled times $t \in [0,1]$. This loss can be justified as a weighted variational lower bound on the data log likelihood under the diffusion model~\citep{kingma2021variational} or as a form of denoising score matching \citep{vincent2011connection, song2019generative}. $w(\lambda_t)$ is a pre-specified weighting function \citep{kingma2021variational}.

\begin{figure*}[t]
\begin{center}\centerline{\includegraphics[width=0.9\linewidth]{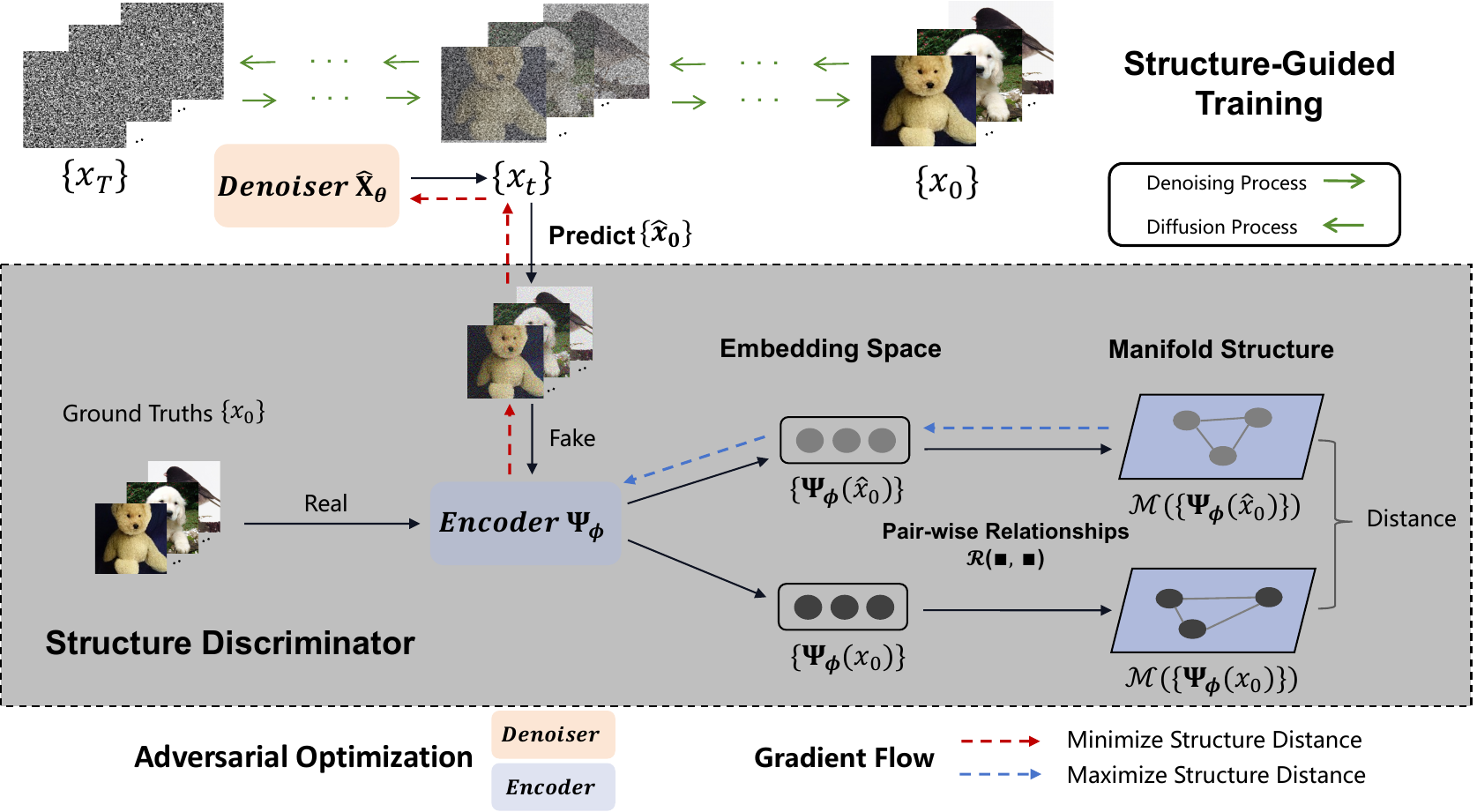}}
\caption{\textbf{Overview of SADM.} We minimize the structural distance between the generated samples (fake) and ground truth samples (real) in the manifold space for optimizing the denoiser, and maximize their structural distance for optimizing the encoder in \textit{structure discriminator}. The denoiser and the structure discriminator are adversarially trained.}
\label{pic-method}
\end{center}
\vspace{-8mm}
\end{figure*}
\section{Proposed Method}
We introduce the proposed \method in detail (\cref{pic-method}). In order to maximally learn the structural information between real samples, we propose structure-guided training of diffusion models in \cref{sec-1}. Then we design a novel \textit{structure discriminator} for adversarially optimizing the training procedure from a structural perspective in \cref{sec-2}. Finally, we alternatively interpret our structure-guided training of diffusion models as \textit{joint sample diffusion} with theoretical analysis for bettwen understanding in \cref{sec-3}.
\subsection{Beyond Instance-Level Training}
\label{sec-1}
We first review the previous instance-level training methods for diffusion models.  
Generally, they train the diffusion model with a finite sample version of \cref{denosing-object-1}:
\begin{equation}
    \mathcal{L}_t = w(\lambda_t)\frac{\sum_{i\in B}||\bx_0^i-\hat\bx_\theta(\bx_{t}^i)||^2}{|B|},
    \label{eq-instance}
\end{equation}
where $\bx_0^i,\bx_{t}^i$ denote the $i^{th}$ ground truth samples and generated samples in the mini-batch $B$ at time step t. However, this objective function encourages the diffusion model to denoise by considering only the instance-level information, neglecting the group-level (structural) information in the mini-batch. Therefore, we enforce the sample predictions of the denoising network to maximally preserve the manifold structures between batch samples.  

\paragraph{Structural Constraint in Manifold} More concretely, ground truth samples $\{\bx^i_0\}_{i=1}^{|B|}$ are first projected from pixel space into embedding space using off-the-shelf pre-trained networks $\Psi(\cdot):\mathbb{R}^{m\times n}\rightarrow \mathbb{R}^{d}$, \eg, an Inception-V3 feature extractor pre-trained on ImageNet \citep{szegedy2016rethinking} (we conduct ablation study in \cref{app-model}). Then, the pair-wise relationships $\mathcal{R}(\Psi(\bx_0^i),\Psi(\bx_0^j))$ are calculated within batch samples $\{\bx^i_0\}_{i=1}^{|B|}$, which contain rich structural information in a low-dimensional manifold. This structural information can be expressed with an affinity matrix $\mathcal{M}(\{\bx^i_0\}_{i=1}^{|B|})$ defined as: 
\begin{equation}
    \begin{bmatrix}
    \mathcal{R}(\Psi_1,\Psi_1) & \mathcal{R}(\Psi_1,\Psi_2)& \cdots   & \mathcal{R}(\Psi_1,\Psi_{|B|})   \\
    \mathcal{R}(\Psi_2,\Psi_1) & \mathcal{R}(\Psi_2,\Psi_2)   & \cdots   & \mathcal{R}(\Psi_2,\Psi_{|B|})   \\
    \vdots & \vdots  & \ddots   & \vdots  \\
    \mathcal{R}(\Psi_{|B|},\Psi_1)  & \mathcal{R}(\Psi_{|B|},\Psi_2)  & \cdots\  & \mathcal{R}(\Psi_{|B|},\Psi_{|B|})  \\
    \end{bmatrix}
\end{equation}
where $\Psi_i$ is a short-hand for $\Psi(\bx_0^i)$, and $\mathcal{R}(\cdot, \cdot)$ denotes the relational function, such as Euclidean distance. Kindly note that many off-the-shelf pre-trained networks are open-source and only work well on clean images, and thus fail to
provide meaningful embeddings when the input is noisy. Therefore, we use the predicted clean samples $\hat \bx_0^i$ for computing affinity matrix, and regularize the denoising network to minimize the structural distance between $\mathcal{M}(\{\bx^i_0\}_{i=1}^{|B|})$ and $\mathcal{M}(\{\hat \bx^i_0\}_{i=1}^{|B|})$. Adding this structural constraint into \cref{eq-instance}, the training objective for denoising network is:
\begin{equation}
    \begin{aligned}
        \mathcal{L}_t &= \frac{\sum_{i\in B}||\bx_0^i-\hat\bx_\theta(\bx_{t}^i)||^2}{|B|}\\
        &+\mathcal{D}(\mathcal{M}(\{\bx^i_0\}_{i=1}^{|B|}), \mathcal{M}(\{\hat \bx^i_0\}_{i=1}^{|B|}))
        \label{denoising-objective-2}
    \end{aligned}
\end{equation}
where $\mathcal{D}(\cdot, \cdot)$ denotes the distance metric between ground truth and predicted affinity matrices. In this way, the diffusion generator (denoiser) is optimized not only to make correct prediction for each instance, but also to preserve the manifold structures of batch samples. 

\subsection{Structure-Guided Adversarial Training}
\label{sec-2}
As demonstrated above, we aim to maximally learn data distribution by aligning the manifold structures of denoiser's predicted samples to those of ground truth samples in each training batch. 
At every training iteration, the manifold structures can be diverse and the denoiser tend to merely focus on some easy-to-learn structures, ultimately leading to trivial solutions that fail to capture the whole data distribution.
In order to mitigate this problem and improve the expressiveness of denoiser, we adversarially learn the denoising network against a \textit{structure discriminator} in a minimax game (\cref{pic-method}), which is trained to distinguish the manifold structures between the real and generated batch samples.

\paragraph{Structure Discriminator} 
Normally, the discriminator in adversarial learning \citep{goodfellow2014generative} would output the discrete value (0 or 1) to determine whether the input is real or fake. However, such discrete discriminator can not be applied in distinguishing manifold structures, because the generated and real sample sets would share similar pair-wise sample relations despite their different feature spaces.
Thus, instead of learning a classification-based discriminator, we design a novel comparison-based structure discriminator to address this issue.

Specifically, as illustrated in \cref{pic-method}, the structure discriminator consists of the aforementioned neural network $\Psi_{\phi}$ with trainable parameters $\phi$ and the distance measure function $\mathcal{D(\cdot,\cdot)}$ that outputs continuous value. 
It projects both real and generated samples into the embedding space, then the structure discriminator is trained against the denoiser for finding a better manifold (or embedding space) to distinguish real sample set from generated set by maximizing their structural distance:
\begin{equation}
        \max_{\phi} \frac{\eta_t\sum_{i,j\in B}
        \mathcal{D}\big(\mathcal{R}(\Psi_{\phi}(\bx^i_0),\Psi_{\phi}(\bx^j_0)), \mathcal{R}(\Psi_{\phi}(\hat\bx_{\theta,t}^i),\Psi_{\phi}(\hat\bx_{\theta,t}^j))),
        }{|B|^2}. \label{eq-structure}
\end{equation}
where we choose $\eta_t=\frac{1}{t}$ as a time-dependent weighting factor. We can use simple measurements for $\mathcal{D}(\cdot,\cdot)$ ($L_2$ distance) and $\mathcal{R}(\cdot,\cdot)$ (cosine similarity), as the critical semantic information has been already encoded by $\Psi_{\phi}$.
Conversely, the denoiser is adversarially optimized for generating more realistic sample set to fool the structure discriminator by minimizing the structural distance. 

\paragraph{Final Optimization Objective} The final training objective of our \method consists of normal denoising score matching loss (\cref{denosing-object-1}) and adversarial structural distance loss (\cref{eq-structure}) at timestep $t$, which can be written as: 
\begin{equation}
    \begin{aligned}
        \mathcal{L}_t(\theta) &= \frac{w(\lambda_t) \sum_{i\in B}\|\hat\bx_\theta(\bx_t^i) - \bx_0^i \|^2_2}{|B|}\\
        +\max_{\phi} & \frac{\sum_{i,j\in B}\eta_t\| \Psi_{\phi}(\bx^i_0)^T\Psi_{\phi}(\bx^j_0)-\Psi_{\phi}(\hat\bx_{\theta,t}^i)^T\Psi_{\phi}(\hat\bx_{\theta,t}^j)\|^2_2}{|B|^2}. \label{objective_final}
    \end{aligned}
\end{equation}
In training, we iteratively optimize the feature extractor and the denoising network in \cref{objective_final}. In an iteration, we first freeze $\theta$ and update $\phi$ by ascending along its gradient 
\begin{equation}
    \nabla_{\phi}\frac{\sum_{i,j\in B}\eta_t\| \Psi_{\phi}(\bx^i_0)^T\Psi_{\phi}(\bx^j_0)-\Psi_{\phi}(\hat\bx_{\theta,t}^i)^T\Psi_{\phi}(\hat\bx_{\theta,t}^j)\|^2_2}{|B|^2},
    \label{eq-adver-phi}
\end{equation}
then we freeze $\phi$ and update $\theta$ by descending along its gradient 
\begin{equation}
    \begin{aligned}
        &\nabla_{\theta}\frac{\sum_{i,j\in B}\eta_t\| \Psi_{\phi}(\bx^i_0)^T\Psi_{\phi}(\bx^j_0)-\Psi_{\phi}(\hat\bx_{\theta,t}^i)^T\Psi_{\phi}(\hat\bx_{\theta,t}^j)\|^2_2}{|B|^2} \\
        &+\nabla_{\theta}\frac{w(\lambda_t) \sum_{i\in B}\|\hat\bx_\theta(\bx_t^i) - \bx_0^i \|^2_2}{|B|}.
        \label{eq-adver-theta}
    \end{aligned}
\end{equation} The proposed training algorithm is presented in \cref{alg: training}. 

\paragraph{Generalizing to Latent Diffusion} Our proposed training algorithm applies not
only to image diffusion but also to latent diffusion, such as LDM \citep{rombach2022high} and LSGM \citep{vahdat2021score}. In this case, the intermediate results
$\bx_t$ are latent codes rather than images. We can use the latent
decoder (\eg, VAE decoder \citep{kingma2013auto}) to project the generated latent codes to images and
then use the same algorithm in the image domain.

\begin{algorithm} 
	\caption{SADM, our proposed algorithm.}
    \label{alg: training}
 	
    \SetKwInOut{Input}{input}\SetKwInOut{Output}{output}

    \Input{$\{\alpha_t\}_{t\in [0,1]},\{\sigma_t\}_{t\in [0,1]}$ the noise schedule, $|B|$ the batch size, $w(\lambda_t),\eta_t$ the scaling factors for denoising score matching loss and adversarial loss, pre-trained encoder $\Psi_{\phi}$ for structure discriminator.}
    Initialize denoising network parameters $\theta$;
    
 	\While{$\theta$ has not converged}{
        Sample a minibatch $\{\bx_0^i\}_{i=1}^{|B|} \sim q_0$ and $\{\bepsilon^j\}_{j=1}^{|B|} \sim \mathcal{N}(\mathbf{0},\mI)$ \\
        Sample $t \sim U[0,1]$\\
        Sample $\{\bx_t^i = \alpha_t\bx_0^i+\sigma_t\bepsilon^i\}_{i=1}^{|B|}$,\\
        Freeze $\phi$ and update $\theta$ with its gradient
        $\nabla_{\theta}\frac{\sum_{i,j\in B}\eta_t\| \Psi_{\phi}(\bx^i_0)^T\Psi_{\phi}(\bx^j_0)-\Psi_{\phi}(\hat\bx_{\theta,t}^i)^T\Psi_{\phi}(\hat\bx_{\theta,t}^j)\|^2_2}{|B|^2} $\\
        $\quad +\nabla_{\theta}\frac{w(\lambda_t) \sum_{i\in B}\|\hat\bx_\theta(\bx_t^i) - \bx_0^i \|^2_2}{|B|}$\\
 	}
  
        Adversarially train denoising network  and structure discriminator by iteratively updating their parameters $\theta,\phi$ according to \cref{eq-adver-phi,eq-adver-theta}.
        
        \Output{Denoising network $\theta$.}
\end{algorithm}
\subsection{Interpreting as Joint Sample Diffusion}
\label{sec-3}
\paragraph{Relation-Conditioned Diffusion Process}
For better understanding of our proposed structure-guided training, we interpret it as a joint sample diffusion model that simultaneously perturbs and denoises a set of samples conditioned on the relation variable. Formally, let $\by_0 = (\bx_0^i,\bx_0^j)$, where $\bx_0^i,\bx_0^j$ are independent random variables sampled from ground truth distribution $q_0$. And the relation variable that encodes the structure information is defined as $ \mathcal{R} = \mathcal{R}(\bx_0^i,\bx_0^j)+ \gamma\bepsilon$, where $\gamma\bepsilon$ denote a small gaussian noise added to the relation to avoid degenerated distribution. Then the forward diffusion jointly perturbs $\by_0$, conditioned on $\mathcal{R}$:
\begin{equation}
    \begin{aligned}
        q(\by_t|\by_0,\mathcal{R}) &= \mathcal{N}(\by_t; \alpha_t \by_0, \sigma_t^2 \bI).
    \end{aligned}
\end{equation}

\begin{table*}[t]
    \centering
    \caption{Quantitative results for class-conditional generation on ImageNet 256$\times$256 and 512$\times$512.}
    \resizebox{0.9\textwidth}{!}{
        \small
        \begin{tabular}{lcccccccc}
            \toprule
            \multirow{2}{*}{Model}& \multicolumn{4}{c}{{ImageNet 256$\times$256}} & \multicolumn{4}{c}{{ImageNet 512$\times$512}}\\
            \cmidrule(lr){2-5} \cmidrule(lr){6-9}
             & {FID} $\downarrow$ & {IS} $\uparrow$ & {Precision} $\uparrow$ & {Recall} $\uparrow$ & {FID} $\downarrow$ & {IS} $\uparrow$ & {Precision} $\uparrow$ & {Recall} $\uparrow$ \\
            \midrule
            BigGAN-deep \citep{brock2018large} & 6.95 & 171.40 & \textbf{0.87} & 0.28  & 8.43 & 177.90 & \textbf{0.88} & 0.29 \\
            StyleGAN-XL \citep{sauer2022stylegan} & 2.30 & 265.12 & 0.78 & 0.53 & {2.41} & 
            {267.75} & 0.77 & 0.52\\
            \midrule
            ADM-G, ADM-U \citep{dhariwal2021diffusion} & 3.94 & 215.84 & 0.83 & 0.53 & 3.85 & 221.72 & 0.84 & 0.53\\
            LDM-4-G  \cite{rombach2022high} & 3.60 & 247.67 & \textbf{0.87} & 0.48 & -& -& -& -\\
            RIN+NoiseSchedule \citep{chen2023importance} & 3.52 & 186.20 & - & - & 3.95 & 216.00 & - & - \\
            SimpleDiffusion \citep{hoogeboom2023simple} & 2.44 & 256.30 & - & - & 3.02 & 248.70 & - & -\\
            DiT-G++ \citep{kim2022refining} & 1.83 & 281.53 & 0.78 & 0.64 &  - & - & - & -\\
            MDT-G \citep{gao2023masked} & 1.79 & 283.01 & 0.81 & 0.61 & - & - & - & - \\
            DiT-XL/2-G \citep{peebles2023scalable} & 2.27 & 278.24 & 0.83 & 0.57 & 3.04 & 240.82 & 0.84 & 0.54\\
             \midrule
             \textbf{DiT-SADM (Ours)} & \textbf{1.58} & \textbf{298.46}& \textbf{0.86} &\textbf{0.66}&\textbf{2.11}&\textbf{251.82}&\textbf{0.87}&\textbf{0.63} \\
            \bottomrule
        \end{tabular}
    }
    \label{table:sota-imagenet}
\end{table*}

To reverse the diffusion process, we need to predict $\by_0$, or equivalently, learn the conditional score function
$\nabla_y\log q_t(y_t|\mathcal{R})$
\citep{dhariwal2021diffusion,ho2022classifier}. This formulation allows us to utilize the auxiliary structural information $\mathcal{R}$ in the sampling process, which usually leads to better performance \cite{dockhorn2021score,singhal2022diffuse}. 
\paragraph{Learning Conditional Score}
To approximate the conditional score function, first we decompose it with Bayes' rule, 
\begin{equation}
    \begin{aligned}
        \nabla_y\log q_t(y_t|\mathcal{R}) &= \nabla_y\log q_t(y_t)+ \nabla_y\log q_t(\mathcal{R}|y_t)\\
        &= \Sigma_{s=i,j}\nabla_x\log q_t(x_t^s)
        +\nabla_y\log q_t(\mathcal{R}|y_t).\label{decompose}
    \end{aligned}
\end{equation} To approximate the first term on the right side of \cref{decompose}, we only need to learn $\hat\bx_{\theta,t}^i,\hat\bx_{\theta,t}^j$, as in standard diffusion models \cite{song2020score}. However, the second term is intractable since $q_t(\mathcal{R}|y_t)$ involves the intractable posterior $q(\by_0|\by_t)$. Note that 
\begin{equation}
    \begin{aligned}
        q_t(\mathcal{R}|y_t) &= \int g(\mathcal{R}|\by_0)q(\by_0|\by_t) d\by_0\\
        &=\E_{\by_0}\left[g(\mathcal{R}|\by_0)|\by_t\right],
    \end{aligned}
\end{equation}
where $g(\mathcal{R}|\by_0)$ is the density function of Gaussian distribution with mean $\mathcal{R}(\bx_0^i,\bx_0^j)$ and variance $\gamma^2$, by the definition of $\mathcal{R}$. As a result, we can use $g(\mathcal{R}|\hat\by_{0,t}) = g(\mathcal{R}|(\hat\bx_{\theta,t}^i,\hat\bx_{\theta,t}^j))$ to approximate $q_t(\mathcal{R}|y_t)$, and train it with $L_2$ loss:
\begin{equation}
    \begin{aligned}
        &\E_{\by_t}\|\E_{\by_0}\left[g(\mathcal{R}|\by_0)|\by_t\right]-g(\mathcal{R}|\hat\by_0(\by_t))\|_2^2\\
        &=\E_{\by_t}\|\E_{\by_0}\left[g(\mathcal{R}|\by_0)-g(\mathcal{R}|\hat\by_0(\by_t))|\by_t\right]\|_2^2\\
        &\leq\E_{\by_0,\by_t}\|g(\mathcal{R}|\by_0)-g(\mathcal{R}|\hat\by_0(\by_t))\|_2^2 \quad (\textit{Jensen Inequality})\\
        &\leq w(\gamma)\E_{\by_0,\by_t}\|\mathcal{R}(\bx_0^i,\bx_0^j)-\mathcal{R}(\hat\bx_{\theta,t}^i,\hat\bx_{\theta,t}^j)\|_2^2 = \mathcal{L}^{\text{structure}}_t,
    \end{aligned}
\end{equation}
where $w(\gamma)$ is a weighting scalar that depends only on $\gamma$. The objective function is the sum of the denoising score matching objective and $\mathcal{L}^{\text{structure}}_t$: 
\begin{equation}
    \begin{aligned}
        &\E_{\bx_0^i,\bepsilon^i,t} w(\lambda_t)\|\hat\bx_\theta(\bx_t^i) - \bx_0^i \|^2_2+\E_{\bx_0^j,\bepsilon^j,t}w(\lambda_t) \|\hat\bx_\theta(\bx_t^j) - \bx_0^j \|^2_2\\
        &+ \Eb{\bx_0^i,\bepsilon^i,\bx_0^j,\bepsilon^j,t} {\mathcal{L}_t^{\text{structure}}},\label{final-object-E}
    \end{aligned}
\end{equation}
which is a variational upper bound of negative log likelihood of the joint sample. Our objective in \cref{denoising-objective-2} can be viewed as a finite-sample version of \cref{final-object-E}, which trains the conditional diffusion model to utilize the structural information. 

\section{Experiments}

\subsection{Image Generation}

\paragraph{Experiment Setup} We experiment on CIFAR-10 \citep{krizhevsky2009learning}, CelebA/FFHQ 64x64 \citep{liu2015deep}, and ImageNet 256x256 \citep{deng2009imagenet}. 
We utilize our SADM to facilitate the training of the diffusion backbones from \citet{karras2022elucidating,vahdat2021score} on CIFAR-10 and FFHQ, from \citet{kim2022soft} on CelebA, and from \citet{peebles2023scalable} (DiT, Diffusion Transformer) on ImageNet. 

\paragraph{Evaluation Metrics} We use Frechet Inception Distance (FID) \citep{heusel2017gans} as the
primary metric for capturing both quality and diversity due to its alignment with human judgement.
We follow the evaluation procedure of ADM \citep{dhariwal2021diffusion} for fair comparisons. For completeness, we also use Inception Score (IS) \citep{salimans2016improved}, Precision and Recall \citep{kynkaanniemi2019improved} as the main metrics
for measuring diversity and distribution coverage.

\paragraph{Implementation Details} We train the denoising network from scratch and use the feature extractor of Inception-V3 \citep{szegedy2016rethinking} pre-trained on ImageNet for initializing the encoder of our structure discriminator. 
At the begining of training, we freeze the pre-trained discriminator encoder and train the denoiser with structure-guided training objective in \cref{denoising-objective-2} until convergence, then we adversarially tune the denoiser and encoder with the objective in \cref{objective_final} for 3 or 4 rounds (500k steps) until they achieve a balance. This training paradigm keeps the same for unconditional and class-conditional generation tasks, and can be easily generalized to score-based diffusion models \citep{song2020score,vahdat2021score} by adding our structural constraint into the final objective functions.
\begin{figure*}[ht]
\begin{center}\centerline{\includegraphics[width=1\linewidth]{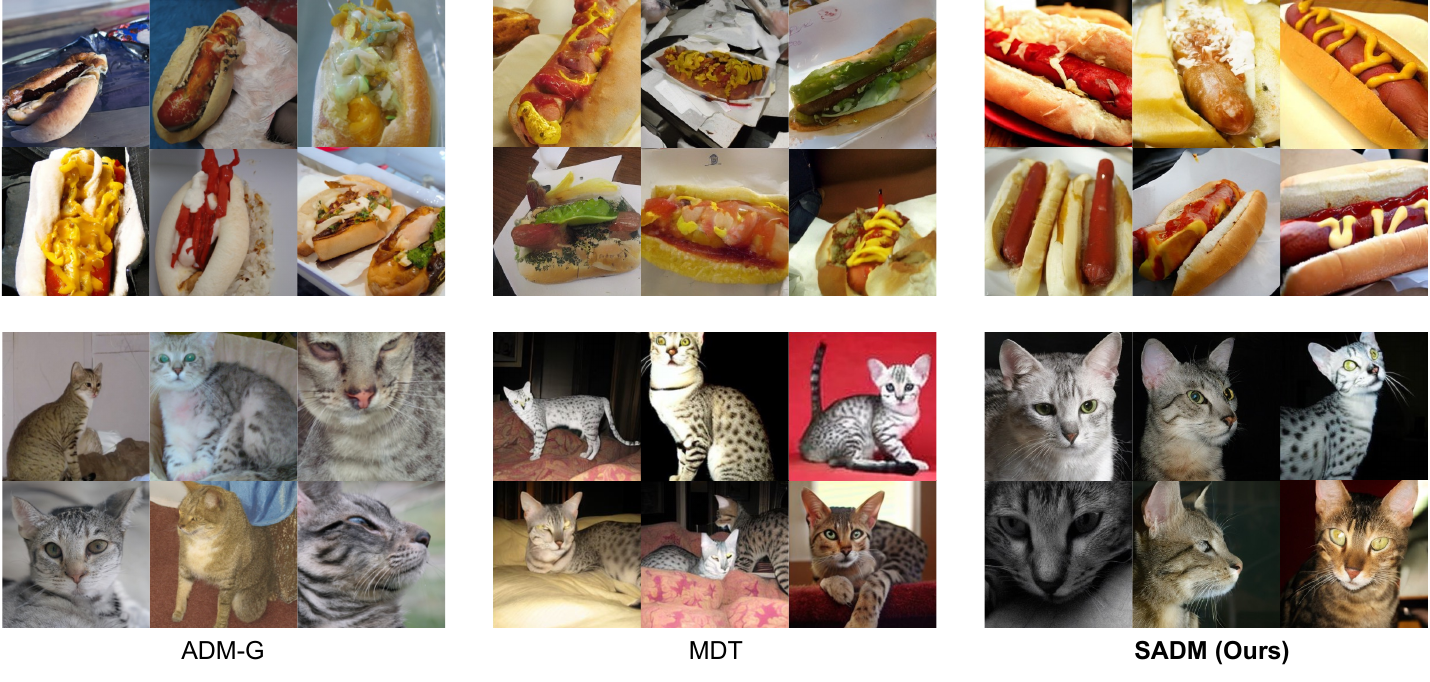}}
\caption{Qualitative comparion with ADM-G \citep{dhariwal2021diffusion} and previous SOTA method MDT \citep{gao2023masked}. Our \method can synthesize more realistic and high-quality samples while maintaining satisfying diversity.}
\label{pic-result2}
\end{center}
\vspace{-8mm}
\end{figure*}

\begin{table}[t]
	\caption{Performance on CIFAR-10.}
	\vskip -0.05in
	\label{tab:cifar10}
	\small
	\centering
	\begin{threeparttable}
    \resizebox{0.48\textwidth}{!}{
		\begin{tabular}{l@{\hskip 0.2cm}c@{\hskip 0.2cm}c@{\hskip 0.3cm}c@{\hskip 0.2cm}c@{\hskip 0.cm}c}
			\toprule
			\multirow{2}{*}{Model} & \multirow{2}{*}{\shortstack{Diffusion \\Space}} & \multirow{2}{*}{NFE$\downarrow$} & \multicolumn{2}{@{\hskip 0.1cm}c@{\hskip 0.8cm}}{Unconditional} & \multicolumn{1}{@{\hskip -0.3cm}c}{Conditional} \\
			& & & NLL$\downarrow$ & FID$\downarrow$ & FID$\downarrow$ \\\midrule
			VDM \cite{kingma2021variational} & Data & 1000 & 2.49 & 7.41 & - \\
			DDPM \cite{ho2020denoising} & Data & 1000 & 3.75 & 3.17 & - \\
			iDDPM \cite{nichol2021improvedddpm} & Data & 1000 & 3.37 & 2.90 & - \\
			Soft Truncation \cite{kim2022soft} & Data & 2000 & 2.91 & 2.47 & - \\
			INDM \cite{kim2022maximum} & Latent & 2000 & 3.09 & 2.28 & - \\
			CLD-SGM \cite{dockhorn2021score} & Data & 312 & 3.31 & 2.25 & - \\
			NCSN++ \cite{song2020score} & Data & 2000 & 3.45 & 2.20 & - \\
			LSGM \cite{vahdat2021score} & Latent & 138 & 3.43 & 2.10 & - \\
			NCSN++-G \cite{chao2022denoising} & Data & 2000 & - & - & 2.25 \\
			EDM \citep{karras2022elucidating} & Data & \textbf{35} & 2.60 & 1.97 & 1.79 \\
			LSGM-G++ \citep{kim2022refining} & Latent &138 & 3.42 & 1.94 & - \\
			EDM-G++ \citep{kim2022refining} & Data & \textbf{35} & 2.55 & 1.77 & 1.64 \\\midrule
            \textbf{SADM} & Latent & 138&2.51&1.78&1.73\\
		  \textbf{SADM} & Data & \textbf{35}&\textbf{2.28}&\textbf{1.54}&\textbf{1.47}\\
			\bottomrule
		\end{tabular}
        }
	\end{threeparttable}
    \label{tab-sota-cifar}
	\vskip -0.1in
\end{table}

\paragraph{Main Results} Our SADM achieves new state-of-the-art FIDs on all datasets including CIFAR-10, CelebA, FFHQ, and ImageNet. 
On ImageNet $256\times 256$ and $512\times 512$, we consistently achieve SOTA FIDs of 1.43 and 2.18 for class-conditional generation as illustrated in \cref{table:sota-imagenet}. Notably, We significantly improve the generation performance of DiT and outperform the previous best FID
of MDT \citep{gao2023masked} solely through improved training algorithm without increasing the model complexity and inference time. 
From \cref{tab-sota-cifar}, we find that our SADM works well for both image diffusion (based on EDM) and
latent diffusion (based on LSGM). 
In experiments, for fair comparisons, we use the same hyperparameters as EDM and LSGM to evaluate the effectiveness of our proposed training algorithm.
And we also achieve significant performance improvement on facial datasets as demonstrated in \cref{tab-sota-celeb}. 
For qualitative results, we compare our \method with ADM-G \citep{dhariwal2021diffusion} and previous SOTA method MDT \citep{gao2023masked} in \cref{pic-result2}.
These remarkable results demonstrate our SADM has the potential for generalizing to arbitrary diffusion architectures and can better learn the whole data distribution.

\begin{table}[t]
\setlength\tabcolsep{2pt}
	\caption{FID performance on CelebA/FFHQ $64\times64$.}
	\vskip -0.05in
	\label{tab:human-face}
	\small
	\centering
	\begin{tabular}{lcccc}
		\toprule
		Model &Diffusion Space& NFE$\downarrow$ & CelebA & FFHQ\\\midrule
		DDPM++ \cite{song2020score} &Data& 131 & 2.32 & - \\
		Soft Truncation \cite{kim2022soft} &Data& 131 & 1.90 & - \\
		Soft Diffusion \cite{daras2022soft} &Data& 300 & 1.85 & - \\
		INDM \cite{kim2022maximum} &Latent& 132 & 1.75 & - \\
		EDM \cite{karras2022elucidating} &Data& 79 & - & 2.39 \\
		Soft Truncation-G++ \citep{kim2022refining} &Data& 131 & 1.34 & - \\
		EDM-G++ \citep{kim2022refining} &Data& 71 & - & 1.98 \\\midrule
        \textbf{SADM}& Latent &131&\textbf{1.28}&\textbf{1.85}\\
        \textbf{SADM}& Data &71&\textbf{1.16}& \textbf{1.71}\\
		\bottomrule
	\end{tabular}
     \label{tab-sota-celeb}
	\vskip -0.1in
\end{table}

\begin{table*}[ht]
	\centering
   \small
 \caption{FID performance comparisons on 8 downstream datasets, all the models are pretrained on ImageNet 256$\times$256.}
	\setlength{\tabcolsep}{4pt}
	\begin{tabular}{l|c|c|c|c|c|c|c|c|c}
			\toprule
			\diagbox{Method}{Dataset}&\makecell[c]{~Food~} & ~SUN~ & \makecell[c]{DF-20M}  & \makecell[c]{Caltech}  & \makecell[c]{CUB-Bird} & \makecell[c]{ArtBench} & \makecell[c]{~Oxford~ \\ Flowers} & \makecell[c]{~Standard~ \\ Cars} &  \makecell[c]{Average \\ FID} \\
            \midrule
			
			AdaptFormer~\cite{adaptformer}    & 13.67 & 11.47 & 22.38 & 35.76 & 7.73 & 38.43  & 21.24 & 10.73 & 20.17 \\
			BitFit~\cite{bitfit}                 & {9.17} & 9.11 & 17.78 & 34.21 & 8.81 & {24.53}   & {20.31} & 10.64 & 16.82 \\
			VPT~\cite{vpt}                  & 18.47 & 14.54 & 32.89 & 42.78 & 17.29 & 40.74 & 25.59 & 22.12 & 26.80  \\
            LoRA ~\cite{lora}                  & 33.75 & 32.53 & 120.25 & 86.05 & 56.03 & 80.99 & 164.13 & 76.24 & 81.25 \\
            DiffFit  \citep{xie2023difffit}            & 6.96 & {8.55} & {17.35} & {33.84} & 5.48 & 20.87   & 20.18 & {9.90} & 15.39 
            \\
            Full Fine-tuning with DDPM                     & 10.46 & {7.96} & {17.26} & 35.25 & {5.68} & 25.31  & 21.05 & {9.79} & {16.59}\\
			\midrule
            \textbf{SADM} (parameter-efficient)&\textbf{5.74}&7.92&16.58&\textbf{32.03}&5.04&\textbf{18.23}&19.37&9.26&{14.27}\\
            \textbf{SADM} (full)&{6.20}&\textbf{7.35}&\textbf{15.12}&32.86&\textbf{4.69}&19.84&\textbf{18.18}&\textbf{8.93}&\textbf{14.15} \\
			\bottomrule
		\end{tabular}
	\label{tab-sota-finetune}
\end{table*}

\subsection{Cross-Domain Fine-Tuning}
\paragraph{Experiment Setup} We conduct cross-domain fine-tuning tasks on diffusion image generation for evaluating the transferability of proposed model, where we pre-train a diffusion model in source domain and adapt it to a target domain by fine-tuning.
Following \citet{xie2023difffit}, we use ImageNet as source dataset, and choose eight commonly-used fine-grained datasets as target datasets: Food101 \citep{bossard2014food}, SUN397 \citep{xiao2010sun},
DF-20M mini \citep{picek2022danish}, Caltech101 \citep{griffin2007caltech}, CUB-200-2011 \citep{Wah2011TheCB}, ArtBench-10 \citep{liao2022artbench},
Oxford Flowers \citep{flowers} and Stanford Cars \citep{cars}. More details about datasets are in \cref{app-imple}.

\paragraph{Implementation Details}
For fair comparison, we follow \citet{xie2023difffit} to set all the hyper-parameters in both pre-training and fine-tuning stages, and use Diffusion Transformer (DiT) \citep{peebles2023scalable} as the diffusion backbone. We pretrain DiT on ImageNet $256\times 256$ with a learning rate of 0.0001 using DDPM objective.
For target datasets, we fine-tune the pre-trained DiT with 24k structure-guide training steps and 4k adversarial training steps. We experiment with two fine-tuning settings, \textit{full} and \textit{parameter-efficient}, for comprehensive evaluations. In parameter-efficient setting, following \citet{xie2023difffit}, we freeze most of parameters in the pre-trained diffusion model and fine-tune the bias term, normalization, and class condition module.

\paragraph{Main Results} 
We achieve SOTA performance on all datasets in fine-tuning tasks as illustrated in \cref{tab-sota-finetune}. Remarkably, we significantly surpass DDPM in full fine-tuning and outperform DiffFit in parameter-efficient fine-tuning.
The results sufficiently demonstrate our superior capability of capturing the whole data distribution, which enables better adaptation to new domains.
Among all datasets, we achieve the best improvement over other methods on ArtBench-10 which has distinct distribution from ImageNet, demonstrating the out-of-distribution generalization ability of our \method. More qualitative results are in \cref{app-samples}.

\subsection{Model Analysis}
\begin{figure}[ht]
\begin{center}\centerline{\includegraphics[width=\linewidth]{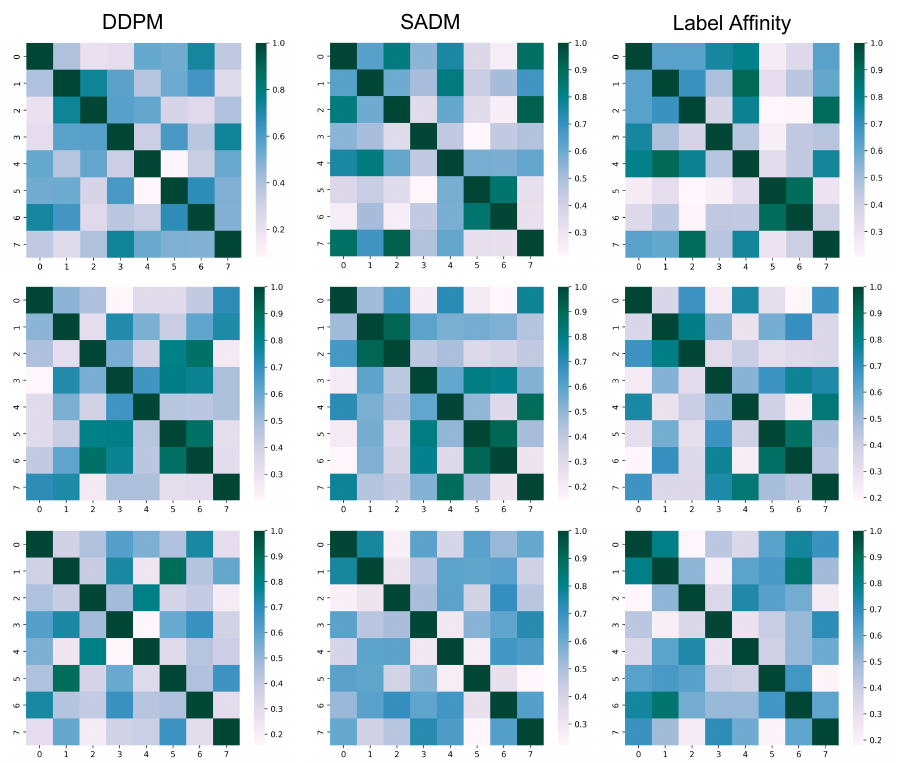}}
\caption{Heatmap visualization with 8 denoised samples.}
\label{pic-heatmap}
\end{center}
\vspace{-8mm}
\end{figure}

\paragraph{Heatmap Analysis}
To evaluate the ability to capture data distribution, we perform heatmap analysis in \cref{pic-heatmap}, where we provide DDPM and our SADM with 8 randomly-selected noisy images in test batch  and visualize the correlations between their denoised outputs. We observe that compared to DDPM, the overall heatmap pattern of our \method is more closer to that of label affinity. The phenomenon demonstrates our \method can precisely learn the manifold structures within real data samples. 
\paragraph{Ablation Study}
We conduct ablation study to validate the effectiveness of our algorithm in \cref{table:ablation}. Here, we base on DDPM \citep{ho2020denoising} architecture, and progressively add our model components (structural guidance and structure discriminator) into it for evaluating FID score on three datasets. We observe that each component can consistently improve the DDPM on all datasets, and the performance improvement of our structural guidance is more significant. The results fully demonstrate the effectiveness of our algorithm. More ablation studies about our model are in \cref{app-model}.

\begin{figure}[ht]
\begin{center}\centerline{\includegraphics[width=0.8\linewidth]{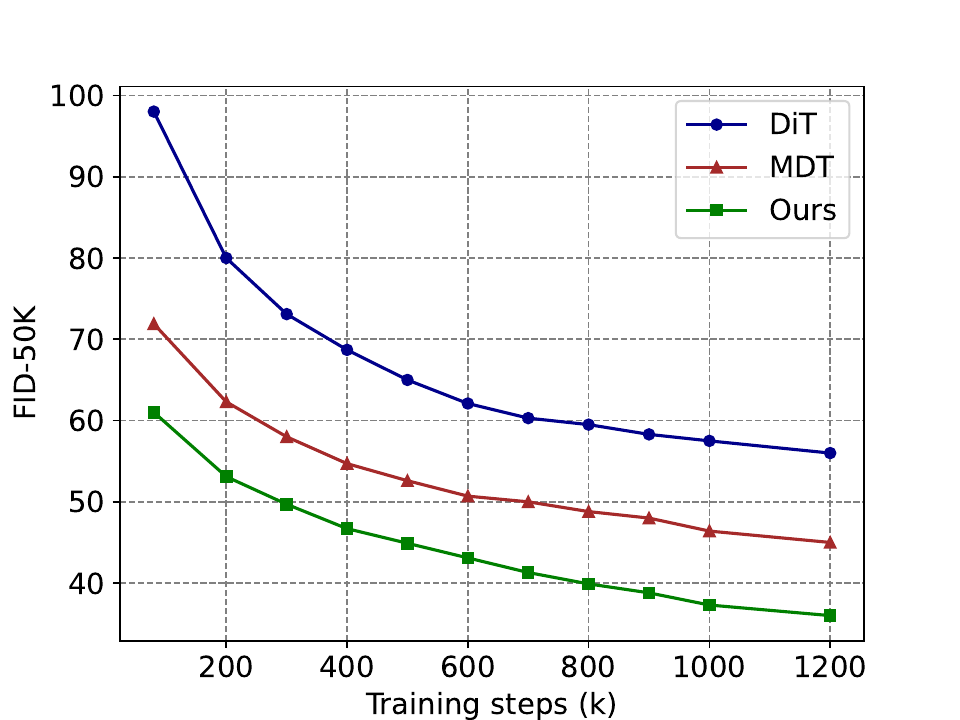}}
\caption{Comparison with SOTA methods on model convergence.}
\label{pic-converge}
\end{center}
\vspace{-8mm}
\end{figure}

\begin{table}[t]
    \centering
    \caption{Ablation study with FID performance. SG denotes structural guidance, SAT denotes SG+Structure Discriminator.}
    \small
        \begin{tabular}{l|c|c|c}
            \toprule
            \diagbox{Dataset}{Model}& {DDPM} & {{+ Our SG}}& + Our SAT \\
            \midrule
            ImageNet&4.59&3.57&3.14\\
            CIFAR-10&3.17&2.64&2.33\\
            CelebA&2.32&1.82&1.64\\
            
            \bottomrule
        \end{tabular}
    \label{table:ablation}
    \vspace{-5mm}
\end{table}
\paragraph{Contributing to Better Convergence}
To investigate the contribution of our structure-guided training to model convergence, we plot the training curve in \cref{pic-converge}. We conclude that compared to previous SOTA methods DiT \citep{peebles2023scalable} and MDT \citep{gao2023masked}, the proposed structure-guided training enables faster and better model convergence because we optimize the diffusion models from a structural perspective, which essentially contributes to capturing the whole data distribution.

\section{Conclusion}
We propose structure-guided adversarial training for optimizing diffusion models from a structural perspective. The proposed training algorithm can easily generalize to both image and latent diffusion models, and consistently improve existing diffusion models with theoretical derivations and empirical results. We achieve new SOTA performance on image generation and cross-domain fine-tuning tasks across 12 image datasets. For future work, we will extend our method to more challenging diffusion-based applications (e.g., text-to-image/video generation).

\section*{Acknowledgement}
This work was supported by the National Natural Science Foundation of
China (No.U23B2048 and U22B2037).
{
    \small
    \bibliographystyle{ieeenat_fullname}
    \bibliography{main}
}

\clearpage
\onecolumn
\appendix
\section{More Implementation Details}
\label{app-imple}
\subsection{Training and Sampling Details}
We present the training and sampling details of our SADM on different datasets in \cref{tab-config} for better reproducing our method. 
\begin{table*}[ht]
	\caption{Training and sampling configurations in SADM.}
	\label{tab:configurations}
	\scriptsize
	\centering
	\begin{tabular}{lccccccc}
		\toprule
		& \multicolumn{2}{c}{CIFAR-10} & \multicolumn{2}{c}{CelebA/FFHQ} & \multicolumn{2}{c}{ImageNet} \\\cmidrule(lr){2-3}\cmidrule(lr){4-5}\cmidrule(lr){6-7}
		& Latent & Image & Latent & Image & Image & Latent \\\midrule
		\multicolumn{7}{l}{\textbf{Training of SADM}}\\
        Based Diffusion Model & LSGM  & EDM & LSGM & EDM & ADM & DiT \\
		Sample Relation Measurement $\mathcal{R}$ &cosine similarity&cosine similarity&cosine similarity&cosine similarity&cosine similarity&cosine similarity\\
        Structural Distance Metric $\mathcal{D}$ &$L_2$ distance&$L_2$ distance&$L_2$ distance&$L_2$ distance&$L_2$ distance&$L_2$ distance\\
        Encoder $\Psi_{\phi}$ of Structure Discriminator &Inception V3&Inception V3&Inception V3&Inception V3&Inception V3&Inception V3\\
        Round of Adversarial Training&2&2&3&3&4&4 \\
        \midrule
		\multicolumn{7}{l}{\textbf{Sampling of SADM}}\\
		SDE & LVP  & WVE & LVP & WVE & LVP & LVP \\

		Solver & PFODE  & PFODE & PFODE & PFODE & DDPM & DDPM \\
		Solver accuracy of $\mathbf{s}_{\bm{\theta}}$ & $1^{\text{st}}$-order & $2^{\text{nd}}$-order & $1^{\text{st}}$-order & $2^{\text{nd}}$-order & $1^{\text{st}}$-order & $1^{\text{st}}$-order \\
		Solver type of $\mathbf{s}_{\bm{\theta}}$ & RK45 & Heun & RK45 & Heun & Euler (DDPM) & Euler (DDPM) \\
		NFE & 138  & 35 & 131 & 71 & 250 & 250 \\
		Classifier Guidance & \xmark & \xmark & \xmark & \xmark & \cmark & \cmark \\
		$w_{t}^{CG}$ & 0  & 0 & 0 & 0 & Adaptive & Adaptive \\
		\bottomrule
	\end{tabular}
 \label{tab-config}
\end{table*}

\subsection{Datasets}
\paragraph{Food101~\cite{bossard2014food}.}
This dataset contains 101 food categories, totaling 101,000 images. Each category includes 750 training images and 250 manually reviewed test images. The training images were kept intentionally uncleaned, preserving some degree of noise, primarily vivid colors and occasionally incorrect labels. All images have been adjusted to a maximum side length of 512 pixels.

\paragraph{SUN 397~\cite{xiao2010sun}.}
The SUN benchmark database comprises 108,753 images labeled into 397 distinct categories. The quantities of images vary among the categories, however, each category is represented by a minimum of 100 images. These images are commonly used in scene understanding applications.

\paragraph{DF20M~\cite{picek2022danish}.}
DF20 is a new fine-grained dataset and benchmark featuring highly accurate class labels based on the taxonomy of observations submitted to the Danish Fungal Atlas. The dataset has a well-defined class hierarchy and a rich observational metadata. It is characterized by a highly imbalanced long-tailed class distribution and a negligible error rate. Importantly, DF20 has no intersection with ImageNet, ensuring unbiased comparison of models fine-tuned from ImageNet checkpoints.

\paragraph{Caltech 101~\cite{griffin2007caltech}.}
The Caltech 101 dataset comprises photos of objects within 101 distinct categories, with roughly 40 to 800 images allocated to each category. The majority of the categories have around 50 images. Each image is approximately 300$\times$200 pixels in size.

\paragraph{CUB-200-2011~\cite{Wah2011TheCB}.}
CUB-200-2011 (Caltech-UCSD Birds-200-2011) is an expansion of the CUB-200 dataset by approximately doubling the number of images per category and adding new annotations for part locations. The dataset consists of 11,788 images divided into 200 categories.

\paragraph{ArtBench-10~\cite{liao2022artbench}.}
ArtBench-10 is a class-balanced, standardized dataset comprising 60,000 high-quality images of artwork annotated with clean and precise labels. It offers several advantages over previous artwork datasets including balanced class distribution, high-quality images, and standardized data collection and pre-processing procedures. It contains 5,000 training images and 1,000 testing images per style.

\paragraph{Oxford Flowers~\cite{flowers}.}
The Oxford 102 Flowers Dataset contains high quality images of 102 commonly occurring flower categories in the United Kingdom. The number of images per category range between 40 and 258. This extensive dataset provides an excellent resource for various computer vision applications, especially those focused on flower recognition and classification.

\paragraph{Stanford Cars~\cite{cars}.}
In the Stanford Cars dataset, there are 16,185 images that display 196 distinct classes of cars. These images are divided into a training and a testing set: 8,144 images for training and 8,041 images for testing. The distribution of samples among classes is almost balanced. Each class represents a specific make, model, and year combination, e.g., the 2012 Tesla Model S or the 2012 BMW M3 coupe.

\section{Ablation Study}
\label{app-model}
In the main text, we have conducted ablation study on our structural guidance and structure discriminator, and find both of them have a critical impact on the final model performance. In this section, we conduct more detailed ablation study on the designs in  structure discriminator for better understanding of our model.
\subsection{Encoder of Structure Discriminator}
\begin{figure*}[ht]
\begin{center}\centerline{\includegraphics[width=0.5\linewidth]{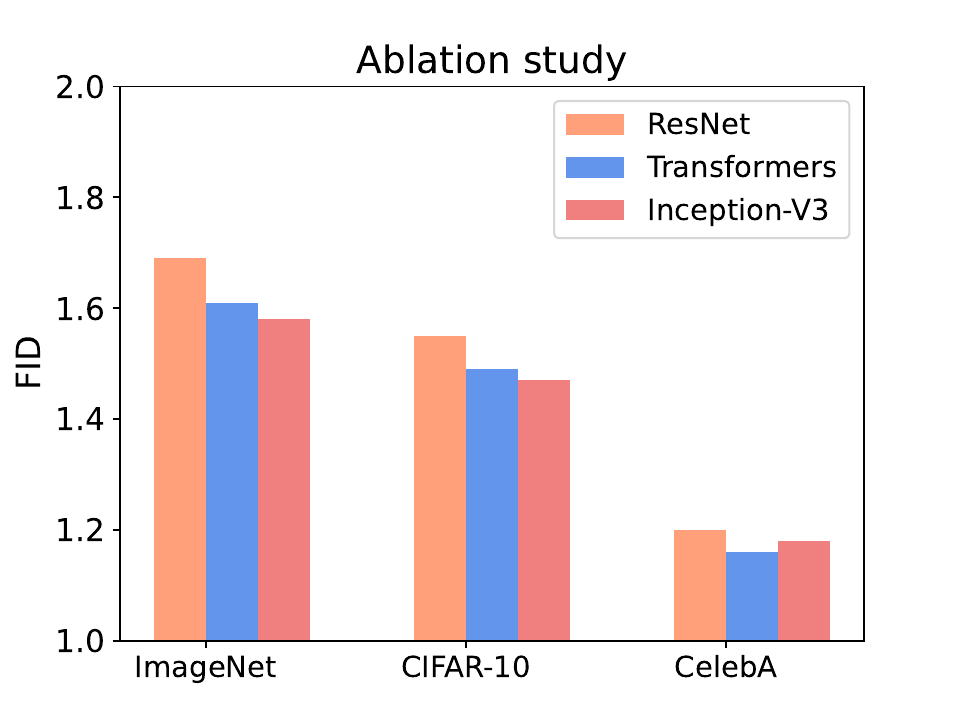}}
\caption{Ablation study on the encoder of structure discriminator in ImageNet, CIFAR-10, and CelebA datasets.}
\label{pic-ablation-encoder}
\end{center}
\end{figure*}

We here conduct ablation study on the encoder choice in our structure discriminator, and we compare with ResNet-18 and Transformer (ViT) architectures that are pre-trained on ImageNet in \cref{pic-ablation-encoder}. 
In the ablation study, we evaluate the FID performance in three datasets with different encoders.
From the results, we can find that Inception and ViT are both better than ResNet-18 because they are superior in capturing the visual semantics of images \citep{wang2023probing,lei2022multi,cheng2023class,li2023vigt}, thus extracting more informative manifold structures. Overall, the encoder choice does not have an obvious impact on the model performance.  

\subsection{Metric of Structure Discriminator}
In main text, we use cosine similarity for $\mathcal{R}$ and $L_2$ distance for $\mathcal{D}$. Here we conduct ablation study on the choice of these metrics, and put the results in \cref{table:ablation-metric}.
In the ablation study, we fix the $\mathcal{R}$ or $\mathcal{D}$ and change the other metric.
We find that using cosine similarity and $L_2$ distance can achieve a similar result, and $L_1$ distance is slightly worse than other metrics. Overall, our model is robust to the choice of metrics.
\begin{table}[ht]
    \centering
    \caption{Ablation study on $\mathcal{R}$ and $\mathcal{D}$ in ImageNet 256$\times$256.}
    
        \begin{tabular}{l|c|c|c}
            \toprule
            \diagbox{Module}{Metric}& $L_1$ distance & $L_2$ distance& cosine similarity \\
            \midrule
            
            Sample Relation $\mathcal{R}$&1.65&\textbf{1.56}&1.58\\
            Structural Distance $\mathcal{D}$&1.63&\textbf{1.58}&1.60\\

            \bottomrule
        \end{tabular}
    \label{table:ablation-metric}
\end{table}
\subsection{Round of Adversarial Training}
\begin{figure*}[ht]
\begin{center}\centerline{\includegraphics[width=0.5\linewidth]{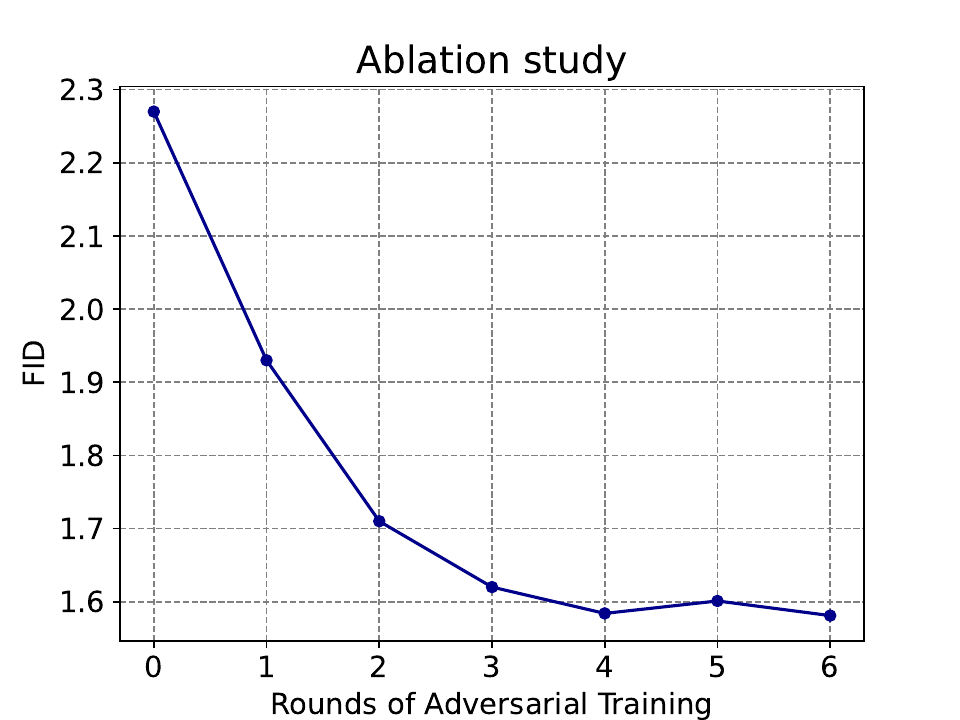}}
\caption{Ablation study on the round of our structure-guided adversarial training in ImageNet.}
\label{pic-ablation-round}
\end{center}
\end{figure*}

We further conduct ablation study on the rounds of our structure-guided adversarial training in \cref{pic-ablation-round}. We find that in the initial round, the model performance can be significantly enhanced regarding FID score, demonstrating the effectiveness of our structure discriminator. After few rounds, the model performance tends to converge as the diffusion denoiser and structure discriminator in \method have achieved a balance.
\section{More Qualitative Comparisons}
\label{app-samples}
\begin{figure*}[ht]
\begin{center}\centerline{\includegraphics[width=0.7\linewidth]{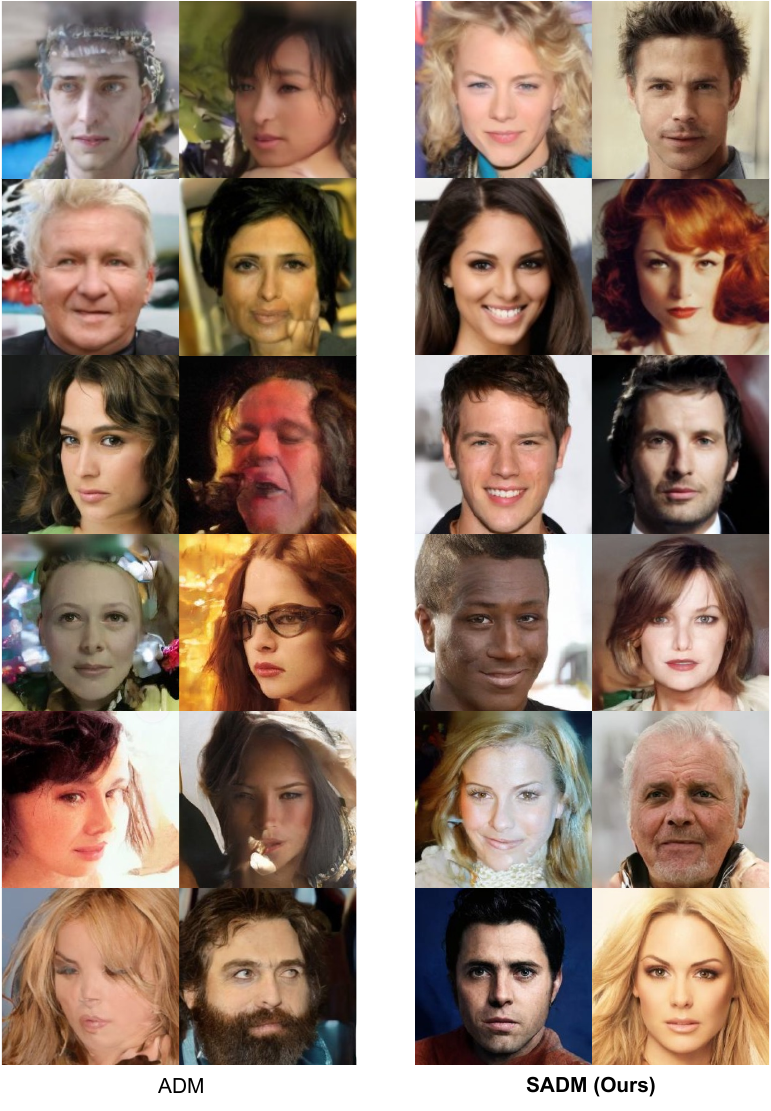}}
\caption{Random generated samples of ADM \citep{dhariwal2021diffusion} and our SADM on unconditional CelebA.}
\label{pic-comp-celeba}
\end{center}
\end{figure*}

\begin{figure*}[ht]
\begin{center}\centerline{\includegraphics[width=0.7\linewidth]{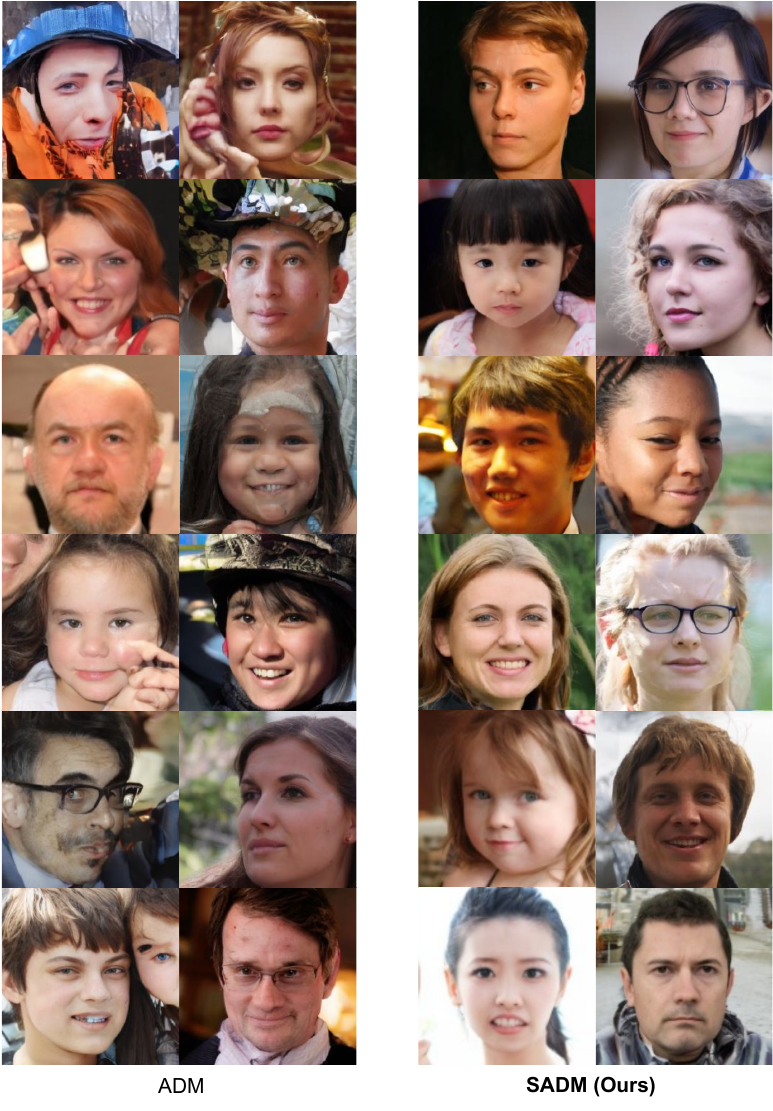}}
\caption{Random generated samples of ADM \citep{dhariwal2021diffusion} and our SADM on unconditional FFHQ.}
\label{pic-comp-ffhq}
\end{center}
\end{figure*}

\begin{figure*}[ht]
\begin{center}\centerline{\includegraphics[width=0.7\linewidth]{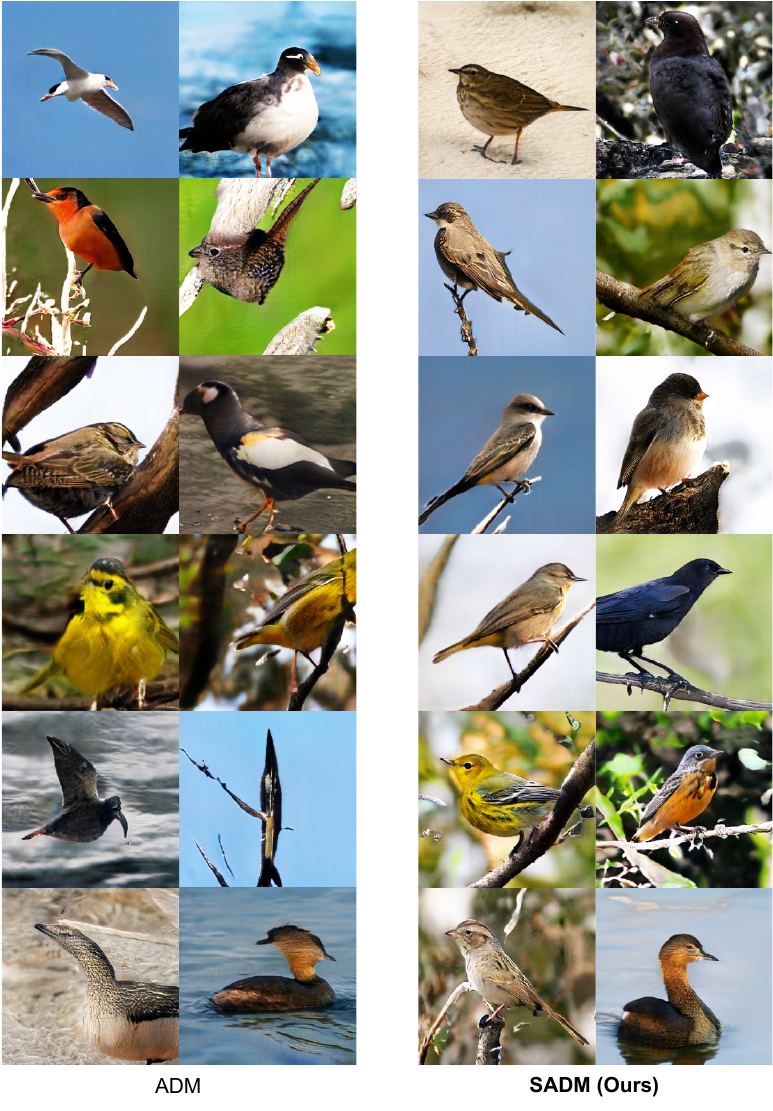}}
\caption{Random generated samples of the diffusion model fine-tuned by ADM \citep{dhariwal2021diffusion} and our SADM on unconditional CUB-200.}
\label{pic-comp-cub}
\end{center}
\end{figure*}

\begin{figure*}[ht]
\begin{center}\centerline{\includegraphics[width=0.7\linewidth]{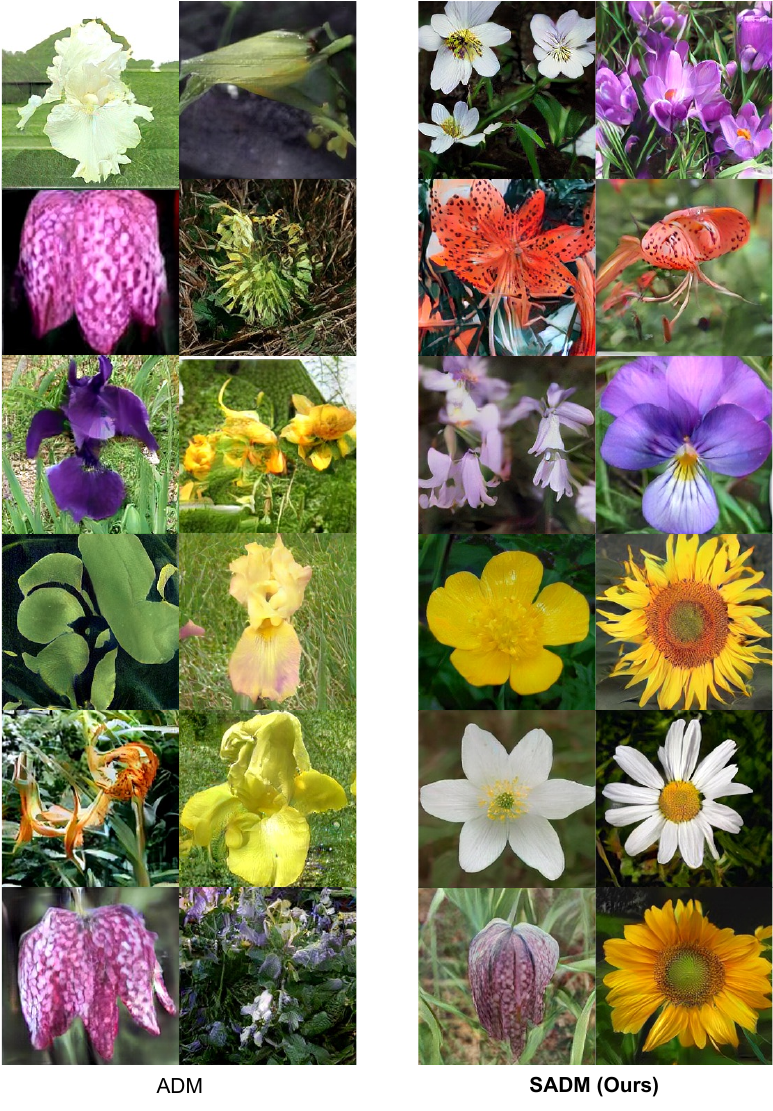}}
\caption{Random generated samples of the diffusion model fine-tuned by ADM \citep{dhariwal2021diffusion} and our SADM on unconditional Oxford-Flowers.}
\label{pic-comp-flower}
\end{center}
\end{figure*}

We here show more qualitative comparison results between our SADM and ADM \citep{dhariwal2021diffusion}. \cref{pic-comp-celeba} and \cref{pic-comp-ffhq} show the generated samples on CelebA and FFHQ datasets in unconditional image generation task, and \cref{pic-comp-cub} and \cref{pic-comp-flower} show the generated samples on CUB-200 and Oxford-Flowers datasets in cross-domain fine-tuning task. We observe that our SADM can comprehensively achieve improvements over previous diffusion models in fidelity and quality, demonstrating the superiority of our new training algorithm. 

\end{document}